\documentclass[journal,twoside, a4paper]{IEEEtran}

\usepackage{color}
\usepackage{amssymb}
\usepackage{rotating}
\usepackage[T1]{fontenc}
\usepackage{lscape}
\usepackage{dblfloatfix}
\usepackage{xcolor}
\usepackage{subcaption}
\usepackage[ruled,vlined]{algorithm2e}

\usepackage[autostyle]{csquotes}
\usepackage{tikz}
\usepackage{acro}
\usepackage{siunitx}
\usepackage{enumitem}
\usepackage{multirow}
\usepackage{cite}
\usepackage{hyperref}
\usepackage{amsmath,amssymb,amsfonts}
\usepackage{tikz}
\usepackage{float}
\usepackage{graphicx}
\usepackage{textcomp}
\usepackage{xcolor}
\usepackage{url}

\usepackage{comment}
\def\BibTeX{{\rm B\kern-.05em{\sc i\kern-.025em b}\kern-.08em
    T\kern-.1667em\lower.7ex\hbox{E}\kern-.125emX}}

\ifCLASSINFOpdf
\else
 
\fi

\usepackage[a4paper, total={180mm,257mm}]{geometry}

\begin{document}

%


\title{Enhancing Kinship Verification through Multiscale Retinex and Combined Deep-Shallow features}




\author{\IEEEauthorblockN{El Ouanas Belabbaci\IEEEauthorrefmark{1},
Mohammed Khammari\IEEEauthorrefmark{2}, 
Ammar Chouchane\IEEEauthorrefmark{3,4},
Mohcene Bessaoudi\IEEEauthorrefmark{4},
Abdelmalik Ouamane\IEEEauthorrefmark{4},
Yassine Himeur\IEEEauthorrefmark{5},
Shadi Atallal\IEEEauthorrefmark{5} and
Wathiq Mansoor\IEEEauthorrefmark{5}
}\\

\IEEEauthorblockA{\IEEEauthorrefmark{1}
LIMED Laboratory, Faculty of Technology , University of Bejaia, 06000 Bejaia, Algeria }\\

\IEEEauthorblockA{\IEEEauthorrefmark{2}
LIMED Laboratory, Faculty of Exact Sciences, University of Bejaia, 06000 Bejaia, Algeria (elouanas.belabbaci@univ-bejaia.dz; mohammed.khammari@univ-bejaia.dz)}\\

\IEEEauthorblockA{\IEEEauthorrefmark{3}University Center of Barika. Amdoukal Road, Barika, 05001, Algeria }\\

\IEEEauthorblockA{\IEEEauthorrefmark{4}dept. Electrical Engineering
LI3C, University of Mohamed Khider Biskra, Algeria
(ouamaneabdealmalik@univ-biskra.dz ; bessaoudi.mohcene@gmail.com)}\\

\IEEEauthorblockA{\IEEEauthorrefmark{5}College of Engineering
and Information Technology
University of Dubai Dubai UAE
(yhimeur@ud.ac.ae)}\\

}

\maketitle

\begin{abstract}
The challenge of kinship verification from facial images represents a cutting-edge and formidable frontier in the realms of pattern recognition and computer vision. This area of study holds a myriad of potential applications, spanning from image annotation and forensic analysis to social media research. Our research stands out by integrating a preprocessing method named Multiscale Retinex (MSR), which elevates image quality and amplifies contrast, ultimately bolstering the end results. Strategically, our methodology capitalizes on the harmonious blend of deep and shallow texture descriptors, merging them proficiently at the score level through the Logistic Regression (LR) method. To elucidate, we employ the Local Phase Quantization (LPQ) descriptor to extract shallow texture characteristics. For deep feature extraction, we turn to the prowess of the VGG16 model, which is pre-trained on a convolutional neural network (CNN). The robustness and efficacy of our method have been put to the test through meticulous experiments on three rigorous kinship datasets, namely: Cornell Kin Face, UB Kin Face, and TS Kin Face.
\end{abstract}


\IEEEpeerreviewmaketitle

\section{Introduction}
Smart cities represent the next phase in urban development, harnessing the power of digital technology, the Internet of Things (IoT), and data analytics to enhance urban life on numerous fronts \cite{atalla2023iot}
Face recognition and biometrics play a pivotal role in the evolution and functionality of smart cities . As urban environments become increasingly interconnected and data-driven, the need for efficient, secure, and personalized services becomes paramount \cite{himeur2023face, chouchane20143d}. Face recognition serves as an advanced tool that aids in public safety, streamlining traffic and crowd management, and even enhancing personalized user experiences in public transportation or retail settings \cite{serraoui2022knowledge}. Concurrently, biometric systems, which go beyond just facial features, offer an added layer of security, ensuring that services are accessed only by authorized individuals. Whether it is for efficient service delivery, security, or fostering a seamless urban experience, face recognition and biometrics are instrumental in realizing the full potential of smart cities, making them safer, more efficient, and responsive to their citizens' needs \cite{himeur2022deep}.

Kinship verification and face recognition are both sub-domains of facial image analysis but serve different primary objectives, albeit with intertwined techniques and methodologies \cite{serraoui2022knowledge, belahcene2016local,ER1 }. 
Kinship verification through facial images aims to ascertain the biological kinship between two individuals by examining their facial characteristics \cite{xia2011kinship,zhang2020advkin, bessaoudi2019multilinear}. 
By verifying kinship ties, cities can ensure that rights related to cultural practices, lands, or hereditary roles are correctly passed to genuine relatives, preserving traditions and heritage.

This image-based verification adds a unique layer to facial analysis by emphasizing the recognition of shared familial traits. This not only adds depth to the challenge of facial image analysis but also broadens its scope \cite{chouchane2023improving, belahcene20142d}. Recognizing kinship is arduous due to the subtle interplay of facial attributes, which include identity, age, gender, ethnicity, and expression \cite{ER1,laiadi2020tensor}. Furthermore, kinship identification has wide-ranging applications. It can be harnessed to organize photos, build family trees, support forensic inquiries, tag images, and aid in locating lost or sought-after individuals \cite{ER2,ER1,bessaoudi2021multilinear}. Though DNA has been the traditional touchstone for verifying kinship, automated facial image algorithms can offer both cost-effective and rapid solutions \cite{zhang2020advkin,fang2010towards,xia2012understanding}.

In this research, we strive to harness the complexities and nuances of these factors to craft a reliable kinship verification system, proficient in overcoming the challenges delineated. To achieve this, we present several innovative contributions and rigorously evaluate our methodology on three renowned datasets: Cornell Kin Face, UB Kin Face, and TS Kin Face. Our primary contributions are outlined as follows:

\begin{itemize}
\item We introduce an advanced preprocessing method termed Multiscale Retinex (MSR). This technique significantly enhances color restoration and overall image quality. Our meticulous experimental evaluation reveals significant improvements in kinship verification outcomes directly linked to the deployment of the MSR approach.

\item For subspace projection and dimensionality reduction, we employ the robust TXQDA+WCCN algorithm, which emphasizes multidimensional data representation.

\item In a bid to refine our feature extraction process, we implement score-level fusion using Logistic Regression (LR). This fusion strategy pairs the shallow texture features of LPQ with the profound features sourced from the VGG16 model. By leveraging the synergies between these attributes, we achieve superior kinship verification performance.

\end{itemize}

The remainder of this article is structured as follows: In Section 2, we discuss related work from three perspectives: Shallow features, CNNs, and Multilinear Subspace Learning for Kinship Verification. Section 3 details our methodology, introducing the preprocessing method (MSR) and the integration of deep and shallow texture features. Section 4 delves into the experimental setup used in our study and discusses the results derived from these experiments. Lastly, in Section 5, we offer concluding remarks that encapsulate the primary findings and contributions of our research.

\section{Related Works}

Over recent years, a multitude of shallow texture models and algorithms for kinship verification have been introduced. These can be broadly categorized into two main streams. The first encompasses methods employing established feature descriptors such as HOG \cite{fang2010towards}, \cite{zhou2016ensemble}, SIFT \cite{lu2013neighborhood}, LBP \cite{lu2013neighborhood}, and D-CBFD as suggested by \cite{yan2019learning}. Typically, these techniques lean on low-level facial features or combinations thereof for kinship verification. The second stream zeroes in on creating straightforward yet distinctive metrics to ascertain if two facial images possess a kinship link. Noteworthy contributions in this realm include the NRML as proposed by Lu et al. \cite{lu2013neighborhood}, PDFL by Yan et al. \cite{yan2014prototype}, and TSL \cite{xia2011kinship}, \cite{xia2012understanding}.

Recently, the Convolutional Neural Network (CNN) has also carved a niche for itself in kinship verification. For example, Li et al. \cite{li2016kinship} put forth the SMCNN, leveraging two identical CNNs supervised by a similarity metric-based loss function. Another innovative technique, termed CNN-points, was introduced by Zhang et al. \cite{zhang122015kinship}. Despite these methodologies showcasing encouraging results, advancements in this domain are somewhat stymied. This is, in part, due to data paucity and the still-evolving understanding of deep convolutional networks.


Multilinear Subspace Learning (MSL) stands as a potent machine learning technique, adept at discerning discriminant features from an array of feature extraction methods, each operating at distinct scales \cite{bessaoudi2019novel,belahcene20143d, ammar_chouchane_146915021}. It is designed to uncover hidden patterns in expansive datasets, making it valuable for discerning relationships amongst various variables. The integration of MSL with tensor data has cemented its stature as a formidable approach for kinship verification endeavors. Among the most pivotal algorithms bolstering kinship verification, \cite{bessaoudi2019multilinear} showcased the Multilinear Side-information-based Discriminant Analysis (MSIDA). MSIDA projects the input region tensor into a novel multilinear subspace. This enhances the separation between samples of different classes while minimizing the distance within samples of the same class. Another notable algorithm in this spectrum is the Tensor Cross-View Quadratic Analysis (TXQDA) \cite{laiadi2020tensor}. TXQDA not only retains the intrinsic data structure and augments the spacing between samples but also adeptly navigates the pitfalls of limited sample sizes, all the while mitigating computational overheads.

\section{Methodology}
\begin{figure*}[t!]
\includegraphics[height=8.8cm, width=19cm]{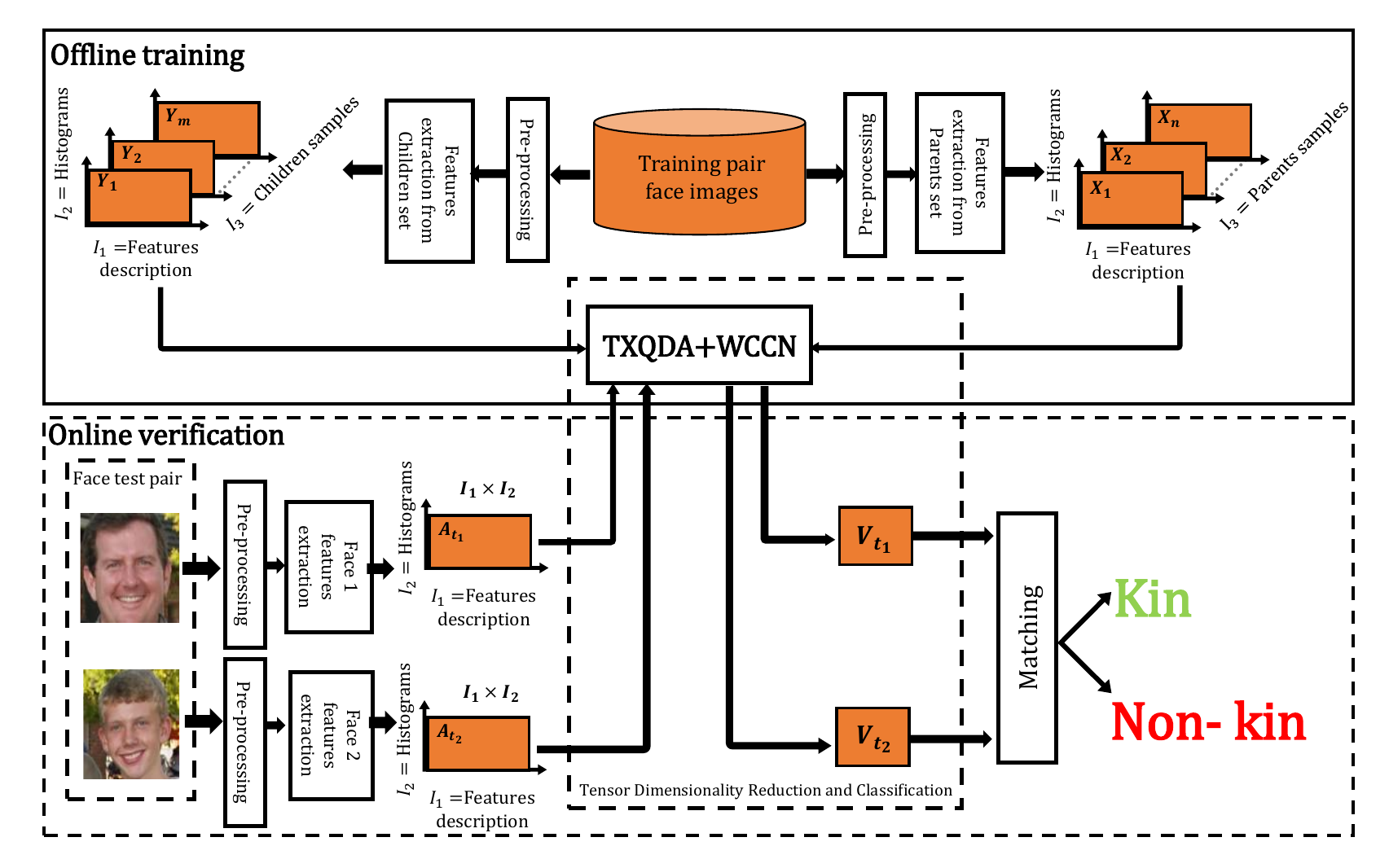}
\caption{Pipeline diagram of the proposed kinship verification and feature extraction using LPQ and VGG16.}
\label{fig1}
\end{figure*}

This section elucidates the architecture of the face kinship verification system posited in our research, as depicted in Fig.\ref{fig1}. The structure encompasses four pivotal components: (A) Face Pre-processing, (B) Feature Extraction, (C) Multilinear Subspace Learning, and (D) Matching and Fusion. We delve into a detailed discourse of each phase in the ensuing sections.

\subsection{Pre-processing} 
\begin{figure}[t!]
 \includegraphics[height=4.5cm, width=9cm]{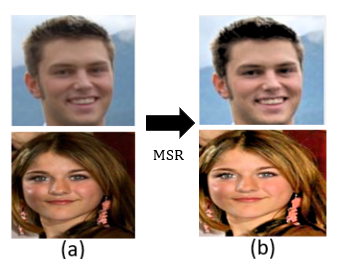}

\caption{Example of  test image processing :(a) original images;(b) MSR images.}
\label{fig2ss}
\end{figure}

In the data pre-processing stage, we employ the MTCNN method \cite{zhang2016joint} to detect facial regions within images. Following this, the MSR algorithm \cite{rahman1996multi} is leveraged for image enhancement. The MSR algorithm amplifies the dynamic range of images while preserving their color accuracy.
Fig. \ref{fig2ss} ilustrates an example of  test image processing :(a) original images;(b) MSR images.

\begin{algorithm}
\caption{Description of the proposed ViT Algorithm.}
\label{alg:ViT}

\textbf{Input:} Image RGB\\
\textbf{Output:} Image MSR\\
\textbf{Initialization:}\\
//\textit{Input Processing}

multiscaleRetinex(image)\\
Split the input image into its RGB channels \\
\textbf{for} each color channel \textbf{do}\\
$L \gets \log(\text{image})$ ~~~~Convert to logarithmic domain\\ \textbf{for} each scale \textbf{do}\\
$F_s \gets \text{SurroundFunction}(L)$
$R_s \gets L - F_s$ 
$W_s \gets R_s \times w_s$ ~~~~Weight the result according to scale
\textbf{endfor}\\
$R \gets \exp\left(\sum_s R_s w_s\right)$ ~~~~~ Sum all the weighted single scale results and convert back to the linear domain\\
\textbf{endfor} \\
Merge all the color channels\\

\end{algorithm}

short for Visual Geometry Group with 16 layers, is a pretrained convolutional neural network (CNN) that is widely acclaimed for its exceptional accuracy in image recognition tasks. It comprises 13 convolutional layers and 3 fully connected layers, specifically the "fc6," "fc7," and "fc8" layers \cite{simonyan2014very}. For shallow feature extraction, we employ the Local Phase Quantization (LPQ) descriptor \cite{ojansivu2008blur}, a well-regarded local texture descriptor. To optimize the verification rate, features are extracted at various scales by adjusting certain parameters, notably the window size $R$.

\subsection{Multilinear subspace learning using TXQDA+WCCN}

In the offline training phase, the TXQDA+WCCN technique involves projecting the training tensors X and Z into a novel discriminant subspace \cite{bessaoudi2021multilinear}. This projection reduces the dimensions of both tensors, resulting in new dimensions $\acute{\mathrm{I}}_1$x$\acute{\mathrm{I}_2}$ for mode-1 and mode-2, respectively, where $\acute{\mathrm{I}}_1$x$\acute{\mathrm{I}_2}$ ${<<}$ $\mathrm{I}_1$x$\mathrm{I}_2$. Notably, the dimension of mode-3 remains consistent, representing the individuals in the dataset. This method aims to reduce the high dimensionality of the higher-order tensor, thereby producing a new feature representation that augments inter-class distinctions while diminishing high intra-class variability.

\subsection{Matching and Logistic Regression Fusion}

Upon projection of the facial image data using the TXQDA+WCCN algorithm, the matching process is executed by computing the cosine distance between two vectors in the discriminant subspace \cite{dehak2010front, chouchane20153d,chouchane2016analyse}. To combine the scores derived from both deep and shallow texture features, we employ a robust technique known as Logistic Regression (LR) \cite{harrell2001regression}. The choice of this fusion technique is influenced by its proven efficacy in prior fusion studies \cite{bessaoudi2021multilinear,BELABBACI2023}. It enables us to harness the advantages of both feature types, leading to enhanced performance in our facial image-matching system.

\section{Experiments}

In this section, we undertake a set of experiments to gauge the efficacy of the proposed kinship verification system. We subject our system to tests using three distinct datasets: Cornell Kin Face, UB Kin Face, and TS Kin Face. The experimental results, gleaned from these datasets, are delineated in Tables \ref{tab2} through \ref{tab5}.

\subsection{Benchmark Datasets}

Cornell Kin Face \cite{fang2010towards}: The dataset being referred to consists of a total of 286 facial images, specifically related to 143 pairs of subjects. The facial images in this dataset depict subjects with a frontal pose and a neutral expression.

UB Kin Face \cite{xia2012understanding}: This dataset consists of 600 images of 400 people, divided into 200 groups of child-young parent (c-yp) and 200 groups of child-old parent (c-op). this dataset is regarded as the first of its kind, presenting a novel approach to the kinship verification problem,  as it includes both young and old face images of parents.

TS Kin Face \cite{qin2015tri}: The Tri-subject kinship face dataset consists of images belonging to the child, mother, and father. The dataset comprises 513 images in the Father, Mother, and Son group, as well as 502 images in the Father, Mother, and Daughter group.

\subsection{Parameter Settings}
In our experiments, we utilize the 5-fold cross-validation protocol \cite{qin2015tri}, \cite{yan2019learning} to evaluate the performance of our approach. This protocol ensures that our results can be directly compared to the state of the art in the field. Prior to analysis, all face images in the datasets undergo pre-processing, specifically the detection of the facial region using the MTCNN method. Additionally, we employ MSR techniques to enhance the quality of the images. Subsequently, we extract two distinct types of features, shallow Texture feature and Deep feature. For shallow texture features, we utilize the LPQ descriptor on the facial image, with a window size of R = {3, 4, 5, 6, 7, 8, and 9}. The facial image is subsequently partitioned into 12 blocks. Each block is summed to create a histogram consisting of 256 bins. These individual histograms are then concatenated, resulting in a final feature vector with a size of (1 × 3072). For deep features, we extract them from the face image with a size of 224 × 224 × 3. We utilize four layers from the VGG16 network, specifically fc6,relu6, fc7, and relu7. The resulting features from these layers are concatenated to form a feature matrix with a size of (4 × 4096).

\subsection{Result analysis and discussion }

\begin{figure}[t!]
 \includegraphics[height=4.8cm, width=9cm]{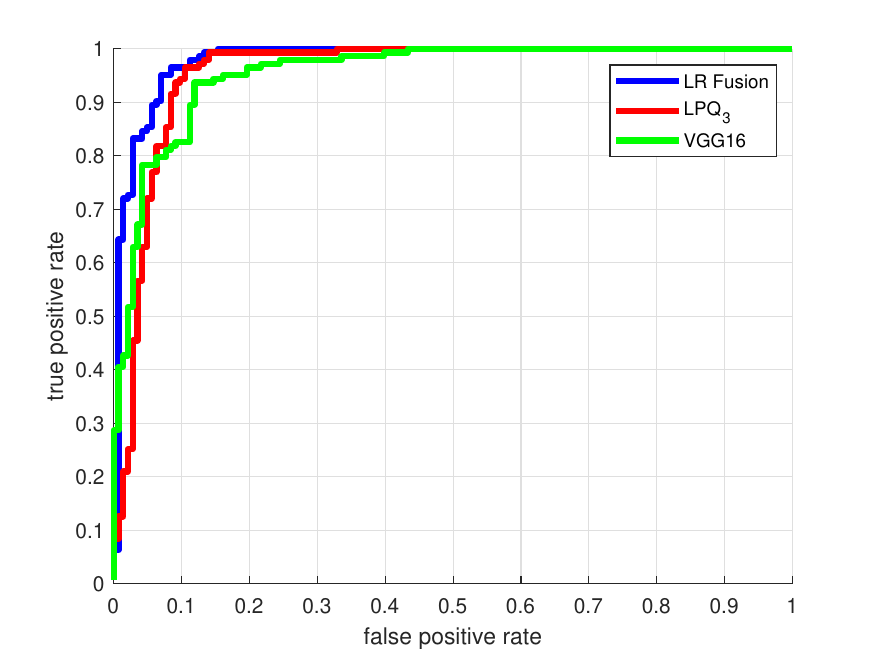}

\caption{ROC curves of the Shallow texture and Deep features on Cornell Dataset.}
\label{fig2C}
\end{figure}

\begin{figure*}
\subfloat[]{\includegraphics[width=9cm,height=5cm]{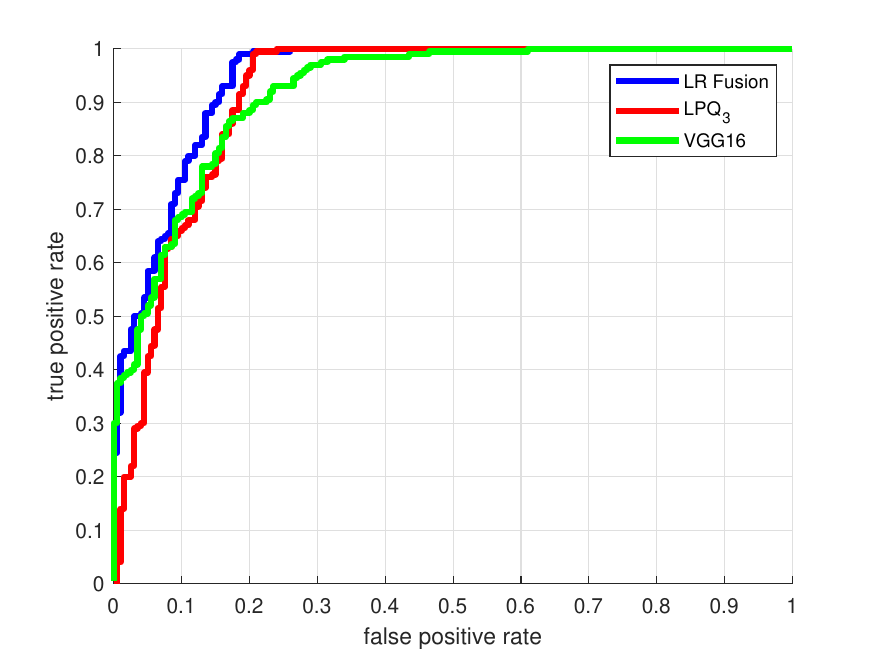}}\hfill
\subfloat[]{\includegraphics[width=9cm,height=5cm]{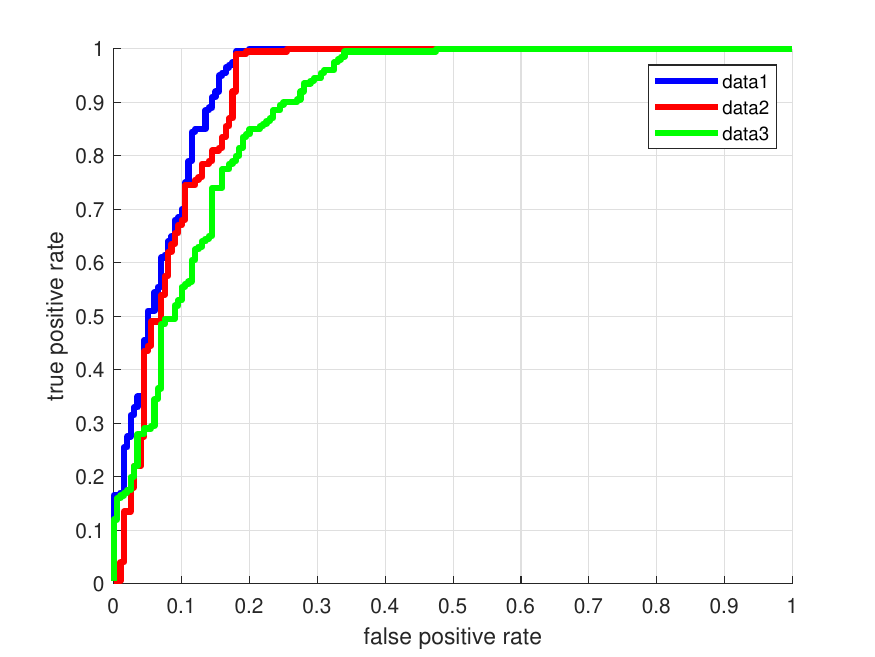}}\hfill
\caption{ROC curves of the Shallow texture and Deep features on UB Kin Face dataset: (a) c-yp set,(b)c-op set.}
\label{fig4u}
\end{figure*}

\begin{figure*}
\subfloat[]{\includegraphics[width=9cm,height=4.9cm]{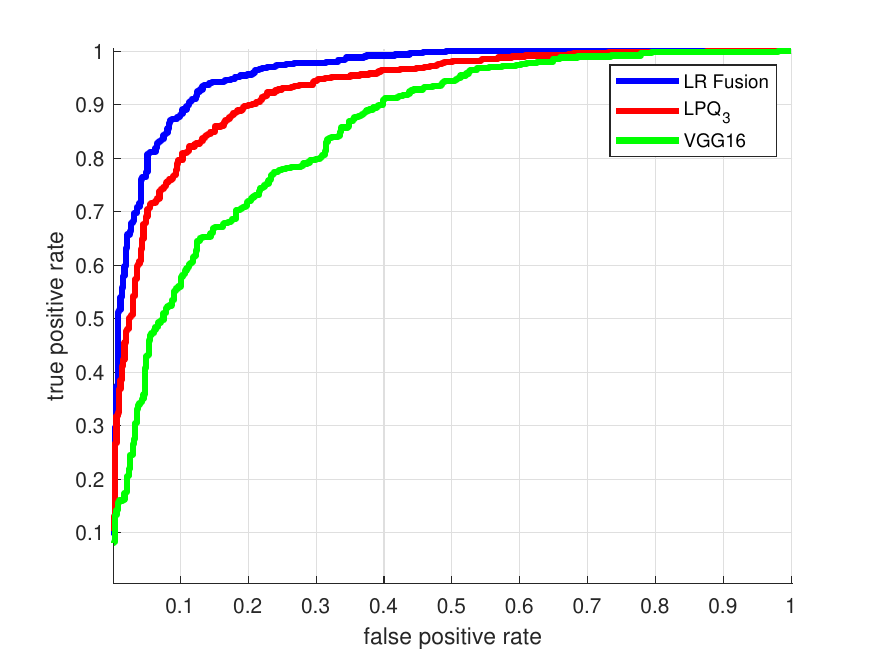}}\hfill
\subfloat[]{\includegraphics[width=9cm,height=4.9cm]{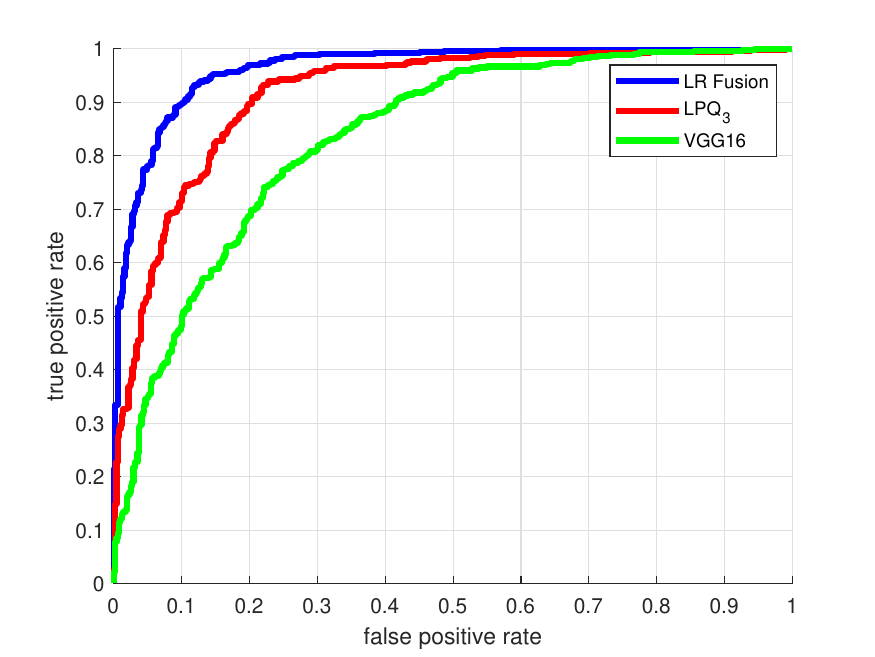}}\hfill
\subfloat[]{\includegraphics[width=9cm,height=4.9cm]{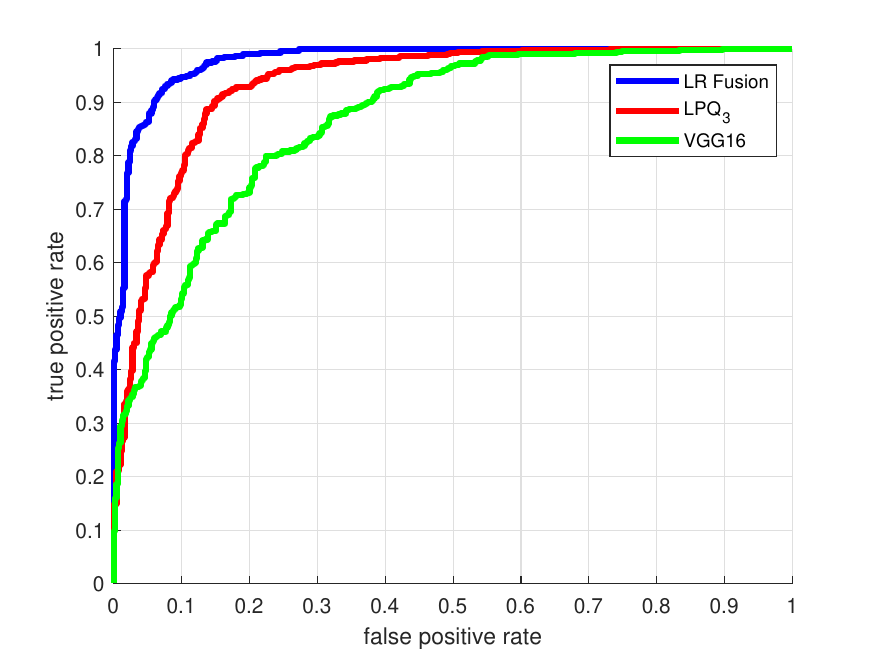}}\hfill
\subfloat[]{\includegraphics[width=9cm,height=4.9cm]{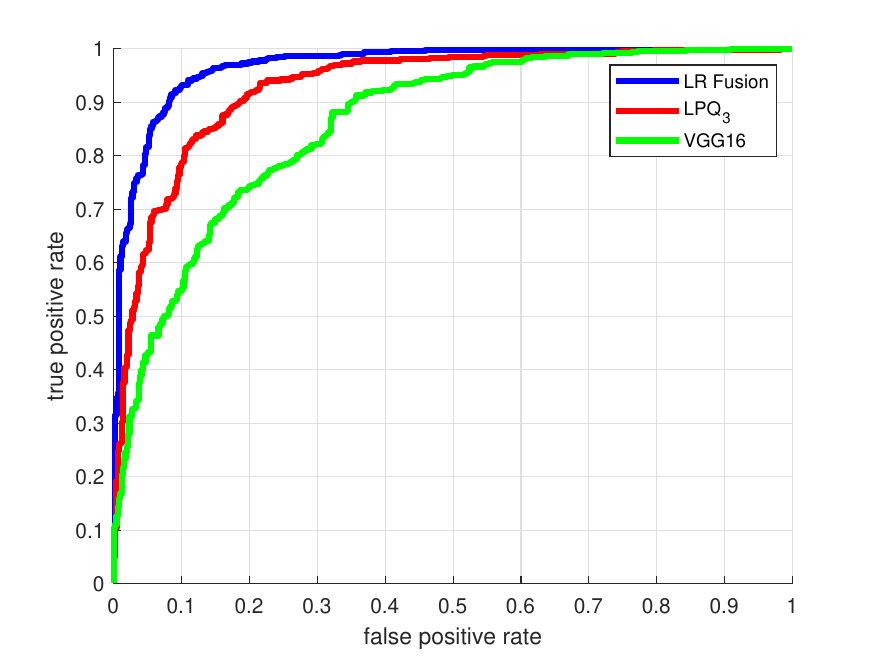}}
\caption{ROC curves of the Shallow texture  and Deep features on TS Kin Face dataset: (a)F–S set, (b) F–D set, (c) M–S set, and (d) M–D.}
\label{fig4t}
\end{figure*}

The experimental results for the Cornell Kin Face, UB Kin Face, and TS Kin Face datasets from our study are presented in Tables I-VI. Tables \ref{tab2}, \ref{tab44}, and \ref{tab4} show the mean accuracy when inputting the original images into our system, with and without a histogram. Additionally, these tables illustrate the outcomes when utilizing LPQ descriptors, both with and without preprocessing. Tables \ref{tab3}, \ref{tab55}, and \ref{tab5} display the mean accuracy of the LPQ descriptor across its 7 scales (R= 3, 4, 5, 6, 7, 8, and 9). These tables also highlight the performance on the fc6, relu6, fc7, and relu7 layers of the pre-trained VGG16 model. Furthermore, they provide details on the average accuracy achieved by fusing the scores from the top-performing LPQ and VGG16 outcomes using LR fusion. Figs. \ref{fig2C}, \ref{fig4u} and \ref{fig4t} depict the ROC curves, showcasing the best results our methodology secured across the three datasets.

\begin{table}[t!]
\caption{The mean accuracy using LPQ with and without (histograms and MSR) on the Cornell Kin Face dataset.}
\begin{center}

\begin{tabular}{|c|c|}
\hline
 \textbf{Settings}  & \textbf{Mean Acc (\%)}      \\
\hline
 Without histogram  & 54.51    \\

  With histogram & 59.34    \\

\hline

 Without MSR  & 76.58    \\

 With MSR &\textbf{ 94.16}    \\
 \hline

\hline
\end{tabular}

\label{tab2}
\end{center}
\end{table}

\begin{table}[t!]
\caption{The mean accuracy using LPQ on Cornell Kin Face dataset with different scales and the LR fusion with VGG16.}
\begin{center}

\begin{tabular}{|c|c|c|}
\hline
 \textbf{Method} & \textbf{Scales}& \textbf{Mean Acc (\%)}        \\
\hline
&  \textbf{R=3}  &      \textbf{94.16}   \\
&  R=4	&     92.72      \\
LPQ& R=5	 &     93.43   \\
&  R=6	 &   93.39     \\
&  R=7 &   92.06      \\
&   R=8  &   93.09     \\
&   R=9  &   92.78       \\
\hline
VGG16 & fc6,relu7, fc7 and relu7 &  91.02       \\

\hline
LR Fusion& 	LPQ (R=3) + VGG16		 &  \textbf{95.18}         \\
	
\hline
\end{tabular}

\label{tab3}
\end{center}
\end{table}

\begin{table}[t!]
\caption{ The mean accuracy using LPQ with and without (histograms and MSR) on the  UB Kin Face dataset.}
\begin{center}
\begin{tabular}{|c|c|c|c|}
\hline
\textbf{Settings}&	\textbf{c-yp}&	\textbf{c-op}&	\textbf{Mean}\\
 &	 &		 & \textbf{Acc (\%)}	\\
\hline
Without histogram &	55.28 &	56.00 &	55.64\\
With histogram    &	60.75&		60.58&	60.67\\
\hline
Without MSR       &	88.17&		88.55&	88.36\\
With MSR          &	\textbf{89.42}&		\textbf{90.78}&	\textbf{90.10}\\

\hline 
\end{tabular}

\label{tab44}
\end{center}
\end{table}

\begin{table}[t!]
\caption{ The mean accuracy using LPQ on UB Kin Face dataset with different scales and the LR fusion with VGG16.}
\begin{center}
\begin{tabular}{|c|c|c|c|c|}
\hline
\textbf{Method} & \textbf{ Scales}&	\textbf{c-yp} &	\textbf{c-op}	& \textbf{Mean}\\
&  &		&  	 &	\textbf{Acc (\%)}	\\
\hline

&\textbf{ R=3 }    &	\textbf{89.42}	    & \textbf{90.78 }&\textbf{	90.10	}\\
& R=4     &	88.92		& 88.29 &	88.61	\\
 LPQ& R=5 & 84.23	   & 86.55 &	85.39	\\
& R=6     & 83.41		& 85.60 &	84.51	\\

& R=7     &	82.43		& 83.36 &	82.90	\\
& R=8     &	80.18		& 82.08 &	81.13	\\
& R=9     & 80.93		& 83.08	&   82.01	\\
\hline 
VGG16& fc6,relu6  & 88.39		& 86.21 &	87.30	\\
 & 		 fc7, and relu7 	 &        & &   \\
\hline 
LR Fusion      & LPQ (R=3) &  \textbf{91.16} &	\textbf{91.52} &	\textbf{91.34}	\\
 &  	+ VGG16 &        & &   \\

\hline 
\end{tabular}

\label{tab55}
\end{center}
\end{table}

\begin{table}[t!]
\caption{ The mean accuracy using LPQ with and without (histograms and MSR) on the  TS Kin Face dataset.}
\begin{center}
\begin{tabular}{|c|c|c|c|c|c|}
\hline
\textbf{Settings}&	\textbf{FS}&	\textbf{FD}&	\textbf{MS}	&\textbf{MD}	&\textbf{Mean}\\
 &	 &	 &	 &	 & \textbf{Acc (\%)}	\\
\hline
Without histogram &	54.46 &	54.26 &	53.76 &	52.97 &	53.86\\
With histogram    &	67.09&	64.13&	66.91&	66.97&	66.27\\
\hline
Without MSR       &	79.41&	77.03&	81.68&	81.98&	80.03\\
With MSR          &\textbf{	85.14}&	\textbf{87.13}&	\textbf{86.83}&	\textbf{88.02}&	\textbf{86.78}\\

\hline 
\end{tabular}

\label{tab4}
\end{center}
\end{table}

\begin{table}[t!]
\caption{ The mean accuracy using LPQ on the TS Kin Face dataset with different scales and the LR fusion with VGG16.}
\begin{center}
\begin{tabular}{|c|c|c|c|c|c|c|}
\hline
\textbf{Method} & \textbf{ Scales}&	\textbf{FS} &	\textbf{FD}&\textbf{MS} & \textbf{MD}	& \textbf{Mean}\\
&  &		&  &	 &	 &	\textbf{Acc (\%)}	\\
\hline

& \textbf{R=3} &	85.94	& \textbf{86.93} &	87.23 &	\textbf{88.22} &	\textbf{87.08}	\\
& R=4 &	\textbf{87.43}	& 86.13	&87.81	& 86.92 &	87.07	\\
 LPQ& R=5 &	86.83	& 86.24	&86.83	& 86.53 &	86.60	\\
& R=6 &  87.41	& 86.03	& 87.91	& 86.52 &	86.97	\\
& R=7 &	85.14	& 85.01	&\textbf{87.97}	& 86.26 &	86.09	\\
& R=8 &	84.93	& 82.89	&84.91	& 85.13 &	84.46	\\
& R=9 & 85.36	& 82.34	&83.56	& 84.50	& 83.94	\\
\hline 
VGG16& fc6,relu6 & 77.38	& 78.12 &	79.32	& 79.70 &	78.63	\\
 & 		 fc7 and relu7 	 &       & & & &   \\
\hline 
LR      & LPQ (R=3) & \textbf{90.30}	& \textbf{91.49}	& \textbf{93.17} &	\textbf{92.28} &	\textbf{91.81}	\\
 Fusion&  	+ VGG16 &       & & & &   \\

\hline 
\end{tabular}

\label{tab5}
\end{center}
\end{table}

\subsection{Discussion }
Based on experiments with our proposed approach, which leverages fusion from two types of features (Deep and Shallow texture), across three datasets (Cornell Kin Face, UB Kin Face, and TS Kin Face), we draw the following conclusions:

\begin{itemize}
\item Integrating the image histogram markedly improves the precision of metric evaluations. The inclusion of the histogram enhanced our system's accuracy by 4.83\% for the Cornell dataset, 5.03\% for the UB dataset, and 12.41\% for the TS dataset. Additionally, the benefits of employing the MSR-based preprocessing technique were evident. This preprocessing step elevated the accuracy rates by 15.43\%, 1.74\%, and 6.75\% on the Cornell, UB, and TS Kin Face datasets, respectively.

\item Our findings underscore the superior performance of LR fusion compared to utilizing individual feature types. In this study, we employed score-level fusion that amalgamates scores generated by the CNN-based VGG16 and the LPQ descriptor. Leveraging the LR fusion method, we realized remarkable accuracy rates: 95.18\% for the Cornell Kin Face dataset, 91.34\% for the UB Kin Face dataset, and 92.81\% for the TS Kin Face dataset. Detailed outcomes can be found in Tables \ref{tab3}, \ref{tab55}, and \ref{tab5}.
\end{itemize}

\subsection{Comparison against the state of the art}
The effectiveness of our proposed method, which involves fusing LPQ and VGG16 scores using the LR fusing technique, is compared to more modern methods in Table \ref{tab7} for the Cornell Kin Face, UB Kin Face, and TS Kin Face datasets. The comparison clearly shows that our proposed technique outperforms the recent state-of-the-art methods on the three datasets.

\begin{table}[t!]
\caption{Performance comparison of kinship verification state of the art on Cornell Kin Face, UB Kin Face, and TS Kin Face datasets}
\begin{center}
\begin{tabular}{|c|c|c|c|c|}
\hline

\textbf{Algorithm}	&     \textbf{  year}&   \textbf{Cornell dataset}	 & \textbf{UB dataset}	 & \textbf{TS dataset} \\
\hline
MSIDA \cite{bessaoudi2019multilinear} & 	2019 &  86.87 &83.34 &85.18\\
\hline
FMRE2 \cite{goyal2021eccentricity} & 	2021& 84.16& 85.03&	90.85\\
\hline
AdvKin \cite{zhang2020advkin}	& 2021		  &  81.40	&75.00 &- \\
\hline
BC2DA \cite{mukherjee2022binary}& 	2022& 83.07	& 83.30& 83.55\\
\hline
TXQEDA \cite{serraoui2022knowledge} & 	2022	& 93.77& - &	90.68 \\
\hline
MLDPL \cite{goyal2023kinship}& 2023&-&87.90 	& 92.40 \\
\hline
\textbf{Proposed}	& \textbf{2023}	& 	\textbf{95.18} & \textbf{91.34}& 	\textbf{91.81} \\

\hline 
\end{tabular}

\label{tab7}
\end{center}
\end{table}

\section{Conclusion}
This study presents a kinship verification system that leverages a novel and efficient facial description method. This method harnesses the power of Logistic Regression (LR) fusion between deep and shallow texture features. Additionally, the system is enhanced with the integration of Multiscale Retinex (MSR), addressing challenges related to contrast, lighting, and noise. This enhancement boosts image quality, ultimately leading to superior performance. Employing tensor subspace learning, our method showcases notable results. The system's efficacy is further amplified by applying LR fusion at the score level of LPQ combined with a pre-trained VGG16. Our results suggest that deep and handcrafted texture attributes synergize effectively at the score level, with the fusion substantially elevating kinship verification accuracy.


\end{document}